\def\year{2021}\relax
\documentclass[letterpaper]{article} 
\usepackage{aaai21}  
\usepackage{times}  
\usepackage{helvet} 
\usepackage{courier}  
\usepackage[hyphens]{url}  
\usepackage{graphicx} 
\usepackage{amsmath,amssymb,amsthm}
\usepackage{dsfont}
\usepackage{natbib}  
\usepackage{caption} 
\DeclareMathOperator*{\argmax}{argmax} 
\usepackage[table]{xcolor}
\usepackage{subcaption}
\usepackage{multirow}
\usepackage{color}

\title{
QVMix and QVMix-Max: Extending the Deep Quality-Value Family of Algorithms to Cooperative Multi-Agent Reinforcement Learning
}
\author {
    Pascal Leroy,\textsuperscript{\rm 1}
    Damien Ernst,\textsuperscript{\rm 1}
    Pierre Geurts,\textsuperscript{\rm 1} \\
    Gilles Louppe,\textsuperscript{\rm 1}
    Jonathan Pisane,\textsuperscript{\rm 2}
    Matthia Sabatelli\textsuperscript{\rm 1} \\
}
\affiliations {
    \textsuperscript{\rm 1} Montefiore Institute, Université de Liège, Belgium \\
    \textsuperscript{\rm 2} John Cockerill Defense, Belgium \\
    pleroy@uliege.be 
    M.Sabatelli@uliege.be
}

\begin{document}

\maketitle

\begin{abstract}
This paper introduces four new algorithms that can be used for tackling multi-agent reinforcement learning (MARL) problems occurring in cooperative settings. All algorithms are based on the Deep Quality-Value (DQV) family of algorithms, a set of techniques that have proven to be successful when dealing with single-agent reinforcement learning problems (SARL). The key idea of DQV algorithms is to jointly learn an approximation of the state-value function $V$, alongside an approximation of the state-action value function $Q$. We follow this principle and generalise these algorithms by introducing two fully decentralised MARL algorithms (IQV and IQV-Max) and two algorithms that are based on the centralised training with decentralised execution training paradigm (QVMix and QVMix-Max). We compare our algorithms with state-of-the-art MARL techniques on the popular StarCraft Multi-Agent Challenge (SMAC) environment. We show competitive results when QVMix and QVMix-Max are compared to well-known MARL techniques such as QMIX and MAVEN and show that QVMix can even outperform them on some of the tested environments, being the algorithm which performs best overall. We hypothesise that this is due to the fact that QVMix suffers less from the overestimation bias of the $Q$ function.
\end{abstract}

\section{Introduction}Reinforcement Learning (RL) is a machine learning paradigm where an agent needs to learn how to interact with an environment \cite{sutton2018reinforcement}. While interacting, the agent performs certain actions, which are all then associated with a certain reward. This reward signal $r$ gives the agent some feedback about the quality of the actions it has performed over time, and the central goal of training an RL agent consists of learning a policy $\pi$, which allows the agent to perform actions which will maximise its expected cumulative reward. Over the years, several progress has been made by combining traditional RL techniques with deep neural networks \cite{lecun2015deep, schmidhuber2015deep}. The successful marriage between RL and deep learning, which comes with the name of deep reinforcement learning (DRL) \cite{franccois2018introduction}, has led to many algorithms that have managed to outperform each other over the years \cite{Mnih2015, hasselt2016deep, hessel2018rainbow, sabatelli2018deepQV}. With the rise of always more successful and powerful algorithms, part of the DRL community has started to shift its attention from single-agent RL (SARL) problems to multi-agent RL (MARL), an RL scenario which is characterised by the presence of multiple agents that interact together. In the special case of cooperative MARL, agents cooperate in order
to achieve a common goal. Cooperation in multi-agent systems is of large practical interest since many real-world problems can be formulated in a cooperative MARL setting. Some examples of these are found in robotics \citep{robot_soccer}, sports \citep{zhao2019multiagent}, and, as we will be approaching in this paper, games.

Sometimes, when compared to more traditional SARL, cooperative MARL can be characterised by the additional challenge of the environment being partially observable. Formally, this corresponds to agents interacting in a decentralised partially observable Markov decision process (Dec-POMDP) \citep{DecPomdp}, which results in agents that only perceive a subset of the state space and that select their actions independently (more details are given in the background section).
It is well known that partial observability can be particularly challenging and prevents a centralised controller to be used to solve these kinds of MARL problems. 
As a result, the MARL community first considered SARL approaches to independently control each agent with a fully decentralised (FD) controller. Later it exploited a method called centralised training with decentralised execution (CTDE), which allows exploiting additional information of the environment.

This paper focuses on these approaches by considering the popular StarCraft Multi-Agent Challenge (SMAC) \cite{samvelyan2019starcraft} as a testbed. We do this by first adapting and then improving the DRL algorithms of the Deep Quality-Value (DQV) family of techniques \cite{sabatelli2020deep}. These algorithms are characterised by jointly learning an approximation of the state-value function and the state-action value function. They have proven to significantly outperform popular algorithms such as DQN \cite{Mnih2015} and DDQN \cite{hasselt2016deep} in SARL. Specifically, this paper has three main contributions:
\begin{enumerate}
    \item We generalise the DQV family of algorithms to cooperative MARL problems and test their performance in an FD training setting.
    \item Based on their overall performance, we identify their fundamental limitations and introduce two novel MARL algorithms, called QVMix and QVMix-Max. Both algorithms combine the original benefits of the DQV algorithms with CTDE, resulting in a better performance than state-of-the-art techniques.
    \item We link the performance of the best-performing algorithm, QVMix, to the overestimation bias of the $Q$ function that characterises model-free RL algorithms. Our experiments suggest that QVMix suffers less from this phenomenon than other algorithms.
\end{enumerate}

The rest of this paper is structured as follows: we first provide further details about the cooperative MARL set-up considered in this paper and briefly present some of the algorithms that have been introduced over the years. We then present our novel algorithm. Then, we define the experimental setup and describe our results. The paper ends with a critical discussion and highlights possible directions for future work.

\section{Background}
\label{sec:background}

This section presents some preliminary knowledge that will be used throughout this paper. We start by formally introducing Dec-POMDPs, since all the experiments that will be reported fall into this specific MARL setting. We then introduce some of the methods that have been introduced over the years for dealing with cooperative MARL and which will serve us as a baseline when investigating the performance of the novel algorithms presented in this paper. 

\subsection{Dec-POMDPs}
In MARL, we can consider Markov Games \cite{MarkovGames}, a framework that is characterised by the fact that each agent has its own reward function. However, when agents cooperate to achieve a common goal and share the reward function, another possible framework is Dec-POMDP \cite{DecPomdp}.
In this paper we formally define a Dec-POMDP by a tuple $[n, \mathcal{S}, O, \mathcal{Z}, \mathcal{U}, R, P, \gamma]$, where $n$ agents cooperate by choosing an action at every timestep $t$.
An agent is denoted by $a \in \{1,...,n\}$.
The agent action space is defined by $\mathcal{U}_a$ such that $\mathcal{U}= \mathcal{U}_1 \times..\times \mathcal{U}_n$ is the combined set of $n$ action spaces, one per agent.
The state of the environment at timestep $t$ is $s_t \in S$ while $o_t^{a} \in \mathcal{Z}$ is the observation of this state perceived by the agent $a$ coming from the observation function $O: \mathcal{S} \times \{1,...,n\} \rightarrow \mathcal{Z}$. At each timestep $t$, each agent simultaneously executes an action $u_t^{a} \in \mathcal{U}_a$ such that the state $s_t$ transits to a new state $s_{t+1}$ with a probability defined by the transition function $P(s_{t+1}|s_t, \mathbf{u_t}): \mathcal{S}^2 \times \mathcal{U} \rightarrow \mathbb{R^+}$ where $\mathbf{u_t} = \bigcup_{a \in \{1,..,n\}} u_t^{a}$. After all agents' actions are performed, a common team reward $r_t = R(s_{t+1}, s_t, \mathbf{u_t}): \mathcal{S}^2 \times \mathcal{U} \rightarrow \mathbb{R}$ is assigned to the agents.
Agent $a$ chooses its action based on its current observation $o_t^{a} \in \mathcal{Z}$ and its history $h_t^{a} \in (\mathcal{Z} \times \mathcal{U}_a)^{t-1}$. 
Its policy is the function $\pi^{a}(u_t^{a}|h_t^{a},o_t^{a}): (\mathcal{Z} \times \mathcal{U}_a)^t \rightarrow \mathbb{R^+}$ that maps, from its history and the current observation at timestep $t$, the probability of taking action $u_t^{a}$ at timestep $t$. 
During an episode, the cumulative discounted reward that is obtained from timestep $t$ over the next $T$ timesteps is defined by $R_{t} = \sum_{i=0}^{T-1} \gamma^i r_{t+i}$, where $\gamma \in [0, 1)$ is the discount factor.
In this paper, the goal of each agent is to maximise its expected $R_0$ over a number of timesteps $T$ defined by the episode time limit. $R_0$ depends on the team's policy defined from the individual agents policies: $\mathbf{\pi}={\pi^1,...,\pi^n}$.
The optimal team's policy $\pi^{*}$ maximises the expected cumulative discounted reward: $\pi^{*} =\argmax_{\pi} R_0$.
To evaluate the team's policy, we consider the state-value function $V^{\pi}(s)=\mathds{E}[R_t | s_t = s, \mathbf{\pi}]$ also called the $V$ function.
Because rewards are shared among agents, the $V$ function has the same value for all agents: $V^{\pi}(s)=V^{\pi^1}(s)=V^{\pi^n}(s)$.
We also define the state-joint-actions value function
$Q^{\mathbf{\pi}}(s,\mathbf{u})=\mathds{E}[R_t | s_t=s, \mathbf{u_t}=\mathbf{u}, \mathbf{\pi}]$, also called $Q$ function, which characterises the expected return of the team when playing action $\mathbf{u}$ at timestep $t$ and following policy $\pi$ afterwards.
The corresponding individual state-action value function of an agent policy is defined by $Q^{\pi^{a}}(s,u)=\mathds{E}[R_t | s_t=s, u^a_t=u, \mathbf{\pi}]=Q^{\mathbf{\pi}}(s,\mathbf{u})$, and denoted later as $Q_a$ for the sake of conciseness.

\subsection{Value-Based Methods}
While several techniques have been introduced over the years, ranging from value-based RL algorithms \citep{Son2019QTRAN:Learning, yang2020qatten, wang2020qplex}, to policy gradient methods \citep{Foerster2017, Du2019LIIRLearning}, in this paper we mostly restrict our analysis to the first family of methods, while keeping a more thorough potential comparison to policy gradient methods as future work.

In value-based RL, the main goal consists of learning \textit{value functions} that indicate how good or bad it is for an agent to select a certain action in a particular state. Formally, this corresponds to learning the optimal $Q$ function which, in a single-agent setting, realises the optimal expected return $Q^{\pi^*}(s,u)= \underset{\pi}{\max}\:Q^{\pi}(s,u) \ \text{for all} \ s\in\mathcal{S} \ \text{and} \ u \in\mathcal{U}.$ Once the optimal $Q$ function has been learned, it becomes a straightforward task to derive an optimal policy by greedy-selection $u^*=\argmax_u Q(s_t, u)$. It is well-known that when the complexity of the environment increases, it is not possible to learn $Q^{\pi^*}(s,u)$ with tabular RL techniques such as Q-Learning \cite{watkins1992q}. To overcome this problem, a function approximator, such as a convolutional neural network parametrised by $\theta$, can be used for modeling $Q(s, u;\theta)\approx Q^{\pi^*}(s, u)$. As first introduced by \citet{Mnih2015} DQN's algorithm, the network can be trained with an objective function that resembles Q-Learning coming in the following form:
\begin{equation}
\begin{split}
    \mathcal{L}(\theta) = \mathds{E}_{\langle .
    \rangle\sim B} \bigg[  
    \big(r_{t} + \gamma \max_{u \in \mathcal{U}} Q(s_{t+1}, u; \theta')
    \\ - Q(s_{t}, u_{t}; \theta)\big)^{2}\bigg]
    \label{eq:DQN_loss}
\end{split}
\end{equation}
where $B$ is the experience replay memory buffer used for storing and uniformly sampling RL trajectories $\langle .
    \rangle$ in the form of $\langle s_{t},a_{t},r_{t},s_{t+1}\rangle$, and 
$\theta'$ 
are the parameters of the target-network. While \citet{Mnih2015} consider 
$\theta$
to be the parameters of a convolutional neural network, when partially observable environments are considered $\theta$ usually parametrises a recurrent neural network (RNN) such as an LSTM \citep{Hochreiter1997LongMemory} or a GRU \citep{Chung2014EmpiricalModeling} as in \citep{Hausknecht2015DeepMDPs}.
Since the agent does not perceive the state $s_t$, the input of such network is the observation $o_t$.
Although Q-Learning \cite{watkins1992q} has inspired the \citet{Mnih2015} DQN algorithm, in what follows we consider its MARL extension instead: Independent Q-Learning (IQL) \citet{Tan1993}.

\subsection{IQL}
IQL consists of a fully decentralised algorithm where each agent learns its respective $Q$ function independently. This is done in a SARL-like fashion with all agents ignoring any type of information other than the one that is provided through their own observations.
Since agents deal with partial observability, it is a common practice to approximate each individual $Q$ function with RNNs.

The problem with IQL is that agents must select actions that maximise $Q(s_t, \mathbf{u_t})$ and not their own $Q$ function. As they cannot know which actions are taken by others, it is usually not possible to approximate this function in a decentralised way.
However, it is possible to approximate $Q(s_t, \mathbf{u_t})$ as a function of individual $Q$ functions such that the taken actions maximise both the joint and the individual $Q$ functions.
If individual $Q$ functions satisfy the Individual-Global-Max condition (IGM) \citep{Son2019QTRAN:Learning}:    $\argmax_{\mathbf{u_t}} Q(s_t, \mathbf{u_t}) =\bigcup_{a}\argmax_{u_t^{a}} Q_{a}(s_t, u_t^{a})$, it is possible to factorise $Q(s_t, \mathbf{u_t})$ with these individual $Q$ functions. This is the core idea defining centralised training in value based methods.
    
\subsection{QMIX} QMIX \citep{Rashid2018} is a CTDE method where the factorisation of the state-joint-actions value function, denoted as $Q_{mix}(s_t, \mathbf{u_t})$, is performed as a monotonic function of the individual $Q$ functions $Q_{mix}=\text{Mixer} \left(Q_{a_1}(s_t, u_t^{a_1}),..,Q_{a_n}(s_t, u_t^{a_n}), s_t\right)$
    where Mixer is a function that ensures IGM through the monotonic $\frac{\partial Q_{mix}(s_t, \mathbf{u_t})}{\partial Q_{a}(s_t, u_t^{a})} \geq 0 \text{ } \forall a \in \{a_1,..,a_n\}$.
    This is satisfied by using a hypernetwork \citep{Ha2016HyperNetworks} defined by a parameter network $h_p(.): \mathcal{S} \rightarrow \mathbb{R}^{|\phi|+}$ which takes the state $s_t$ as input and predicts the strictly positive parameters\footnote{To be exact, the offset defined by $h_p()$ is not constrained to be positive.} $\phi$ of the main network $h_o(.): \mathbb{R}^n \times \phi \rightarrow \mathbb{R}$, called the mixer network, therefore, $Q_{mix}(s_t, \mathbf{u_t}) = h_o\left(Q_{a_1}(),..,Q_{a_n}(), h_p(s_t)\right)$.
    The monotonicity of $Q_{mix}$ w.r.t the individual $Q$ functions is satisfied because a neural network made of monotonic functions and strictly positive weights is monotonic w.r.t to its inputs. In addition, the outputs of the $Q$-Mixer depend on the state thanks to the hypernetwork.
    The optimisation procedure follows the same principles used by the DQN algorithm but applied to $Q_{mix}(s_t, \mathbf{u_t})$.
    
\subsection{MAVEN}
    \citet{Mahajan2019MAVEN:Exploration} demonstrated that in certain circumstances, the exploration capabilities of QMIX are limited. 
    To tackle this problem, they added a latent space to influence the behaviour of the agents. It does so by being the input of a parameter network that computes parameters of the fully connected layer linking recurrent cells to the output in the individual $Q$ network.
    This latent variable $z$ is generated once an episode begins by a hierarchical policy network, taking as input the initial state of the environment together with a random variable which is typically discrete and sampled from a uniform distribution.
    The objective that is used for training MAVEN's networks is composed of three parts.
    First, once the latent variable network is fixed the two hypernetworks (the mixer and the latent space parameter network) and the recurrent networks are trained using the same loss that is also used by QMIX. Second, when these three networks are fixed, the latent variable network can be trained using any policy optimisation such as the policy gradient method \citep{NIPS1999_464d828b}, computed with the total sum of rewards per episode. Finally, the influence of the latent variable on the agents behaviour is enforced via a mutual information loss between the latent variable and consecutive transitions. For more details on how MAVEN works we refer the reader to \citep{Mahajan2019MAVEN:Exploration}.

\section{Generalising the DQV family of algorithms to MARL}
\label{sec:qv_mix_family}

Our contributions build on top of the work revolving around the Deep Quality-Value (DQV) family of DRL algorithms \cite{sabatelli2018deepQV, sabatelli2020deep}. The core idea of these SARL techniques is to learn an approximation of the state-value function $V^{\pi}(s)=\mathds{E}[\sum_{k=0}^{\infty}\gamma^{k}r_{t+k} | s_t = s, \pi ]$ alongside an approximation of the state-action value function $Q^{\pi}(s,u)=\mathds{E}[\sum_{k=0}^{\infty}\gamma^{k}r_{t+k} | s_t = s, u_t=u, \pi]$.
The DQV algorithms generalise the tabular RL algorithms presented in \cite{wiering2005qv} and \cite{wiering2009qv} and use deep neural networks serving as value function approximators. Learning a joint approximation of the $V$ function and the $Q$ function can be done in two different ways, both of them requiring two different neural networks for learning $Q(s, u;\theta)\approx Q^{\pi^*}(s, u)$ and $V(s;\phi)\approx V^{\pi^*}(s)$.

\begin{itemize}
    \item\textbf{DQV-Learning} learns an approximation of the state-value function by minimising the following objective
    \begin{equation}
    \begin{split}
        \mathcal{L}(\phi) = \mathds{E}_{\langle 
        .
        \rangle\sim B} \bigg[\big(r_{t} + \gamma V(s_{t+1}; \phi')\\ - V(s_{t}; \phi)\big)^{2}\bigg],
        \label{eq:dqv_v_update}
    \end{split}
    \end{equation}
    while the following loss is minimised for learning the $Q$ function:
    \begin{equation}
    \begin{split}
        \mathcal{L}(\theta) = \mathds{E}_{\langle .
        \rangle\sim B} \bigg[\big(r_{t} + \gamma V(s_{t+1}; \phi') \\ - Q(s_{t}, u_{t}; \theta)\big)^{2}\bigg]. 
    \label{eq:dqv_q_update}
    \end{split}
    \end{equation}
    \item \textbf{DQV-Max Learning} uses the same equation presented in Eq. \eqref{eq:dqv_q_update} for learning the $Q$ function and the following loss for learning the $V$ function:
    \begin{equation}
        \begin{split}
            \mathcal{L}(\phi) = \mathds{E}_{\langle .
            \rangle\sim B} \bigg[\big(r_{t} + \gamma \: \underset{u\in \mathcal{U}}{\max}\: Q(s_{t+1}, u; \theta') \\ - V(s_{t}; \phi)\big)^{2}\bigg].
            \label{eq:dqv_max_v}
        \end{split}    
    \end{equation}
\end{itemize}

We test the update rules of DQV and DQV-Max in a FD training setting and in a CTDE setting. In the first case DQV and DQV-Max follow the same principles as the ones of Independent Q-Learning and we therefore refer to these algorithms as IQV and IQV-Max. In the second case, both algorithms are used in combination with hypernetworks in a QMIX like fashion. We refer to these algorithms as QVMix and QVMix-Max.

\section{Experimental setup}
\label{sec:exp_setup}
\begin{figure}
     \centering
     \begin{subfigure}[b]{0.45\linewidth}
         \centering
         \includegraphics[height=.5\linewidth]{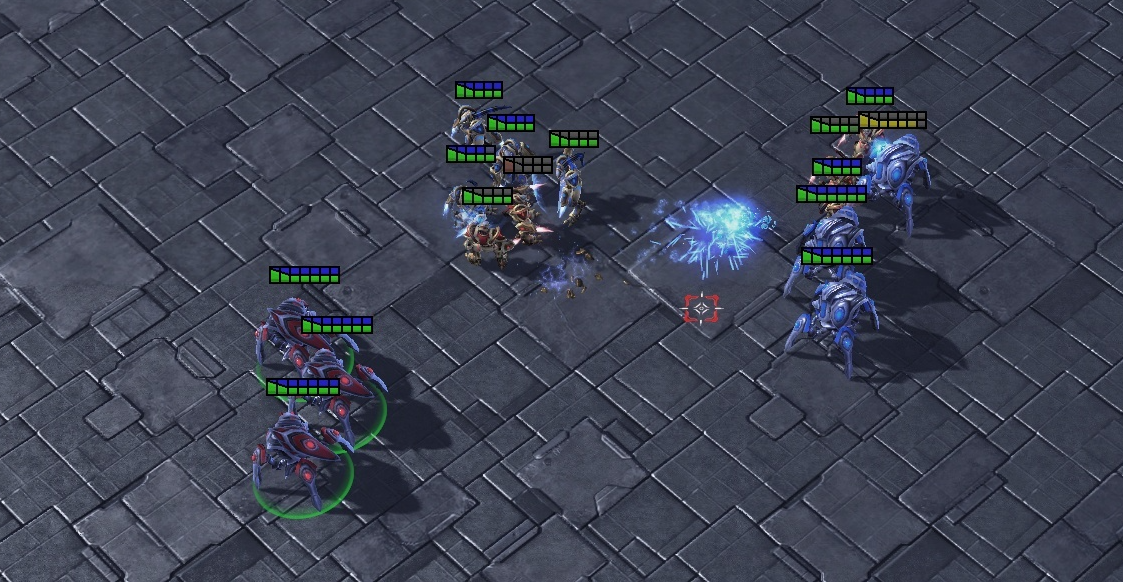}
         \caption{\texttt{3s5z}}
         \label{fig:3s5zsc2}
     \end{subfigure}
     \begin{subfigure}[b]{0.45\linewidth}
         \centering
         \includegraphics[height=.5\linewidth]{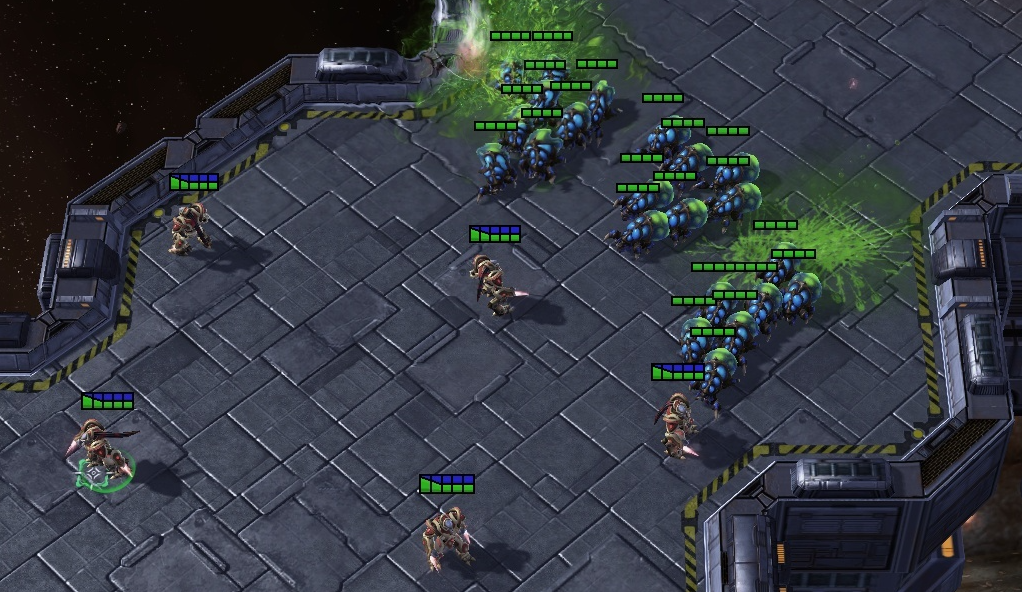}
         \caption{\texttt{so\_many\_baneling}}
         \label{fig:somanybane}
     \end{subfigure}
    \caption{Two pictures of different SMAC scenarios:\texttt{3s5z} at the top and \texttt{so\_many\_baneling} at the bottom.}
    \label{fig:sc2_example}
\end{figure}
We test a total of seven algorithms: the four DQV extensions introduced in the previous section are compared to the three algorithms introduced in the background section: IQL, QMIX and MAVEN. As a test-bed, we use the popular SMAC \cite{samvelyan2019starcraft} environment and evaluate the methods on eight different scenarios (also called maps): \texttt{3m}, \texttt{8m}, \texttt{so\_many\_baneling}, \texttt{2m\_vs\_1z}, \texttt{MMM}, \texttt{2s3z}, \texttt{3s5z} and \texttt{3s\_vs\_3z}. We refer the reader to \citet{samvelyan2019starcraft} for a more in-depth explanation of the differences between map configurations. It is worth noting that the maps chosen for our experiments differ in terms of complexity. Two examples of different scenarios are visually represented in Fig. \ref{fig:sc2_example}. No matter which map is used, the SMAC environment provides the different agents with the following information: agents can observe their health, type, last action performed and a list of possible directions they can move to. If other agents are within their sight range, they can also observe which unit type these agents are part of, how distant they are, and their health. More importantly, if other agents are an agent's ally, their last performed action is visible, if not, the only information provided corresponds to whether such an agent is within an attack range. All this information is encoded as a Boolean or a value $\in [0,1]$. 
Among the different possible actions, agents can decide whether to move in one of the four cardinal directions, not to perform any action at all, or attack an opponent if the latter one is within their attack range. Each attack has the effect of removing a certain amount of health from the opponent. At each time step, the agent receives a zero or positive reward, shared among the team's agents. The goal is to maximise the reward, achieved by reducing each opponent team unit's health to zero, which is then called a win.

Each algorithm is trained on every map ten times, where each neural network is trained from scratch. We train the network for either $5m$ or $10m$ timesteps depending on the considered map and the time that is required to achieve convergence. Every $20000$ timesteps, the parameters of the networks are saved and we perform $24$ testing episodes.

To keep the comparison among the different tested algorithms as fair as possible, we ensure that each algorithm uses the same type of hyper-parameters ranging. Specifically, we referred to the authors of QMIX, MAVEN and IQL to determine the set of hyper-parameters and have simply kept the same values for QVMix, QVMix-Max, IQV and IQV-Max. For a more thorough presentation of all used hyper-parameters we refer the reader to the open-sourced code\footnote{\url{https://github.com/PaLeroy/QVMix}}.
As is common practice within the literature, to improve the learning speed of the algorithms, the parameters of the individual networks are shared among agents.  

\section{Results}
\label{sec:results}

\begin{table*}[h]
    \centering
    \begin{tabular}{|c|c|c|c|c|c||c|c|c|}
        \hline
        
        Training steps & Map & QMIX & MAVEN &     QVMix & QVMix-Max  & IQL & IQV & IQV-Max\\
        \hline
        \multirow{4}{*}{$5m$}  & \texttt{3m} & \cellcolor{yellow!25}100 & 98.7 &   \cellcolor{green!25}100 & 100 & 93.3 & 93.3& 96.6 \\
        
        & \texttt{8m} & 96.6& \cellcolor{yellow!25}98.3 &  \cellcolor{green!25}100 & 96.6 & 83.3& 93.3& 90\\
        
        & \texttt{so\_many\_baneling} & \cellcolor{yellow!25}100 & 97  & \cellcolor{green!25}100& 100& 50 & 40 & 40\\
        
        & \texttt{2m\_vs\_1z} & \cellcolor{yellow!25}100 & 100 &\cellcolor{green!25}100 & 96.6  & 100 & 100 & 100\\
        \hline
        
       \multirow{4}{*}{$10m$} &\texttt{MMM} & \cellcolor{green!25}100 & \cellcolor{yellow!25}97.0 & 93.3 & 96.6  & 61.6 & 83.3 & 50 \\
        
          &\texttt{2s3z}  & 96.6 & \cellcolor{yellow!25}97.5 &  96.6 & \cellcolor{green!25}100 & 59.9 & 56.6 & 40 \\
        
         &\texttt{3s5z} & 40 & 40.8 &\cellcolor{green!25}86.6 & \cellcolor{yellow!25}43.3  & 16.6 & 13.3 & 0 \\
        
         &\texttt{3s\_vs\_3z}  & \cellcolor{yellow!25}100 & 97.9  &  \cellcolor{green!25}100 & 100  & 83.3 & 76.6 & 63.3 \\
        \hline
    \end{tabular}
    \caption{
    Means of win-rates achieved at the end of training by QMIX, MAVEN, QVMix, QVMix-Max, IQL, IQV, and IQVMax in eight scenarios. 
    In the first four scenarios, \texttt{3m}, \texttt{8m}, \texttt{so\_many\_baneling} and \texttt{2m\_vs\_1z}, it is measured after $5$ millions training timesteps.
    In the last four, \texttt{MMM}, \texttt{2s3z}, \texttt{3s5z} and \texttt{3s\_vs\_3z} it is measured after $10$ millions training timesteps.
    In the green and yellow cells, we report the best and second-best means respectively. When results are equivalent, the cells report the fastest and second-fastest method that reaches a win-rate of $100\%$ as it can be seen in Figure \ref{fig:all_win_curves}.}
    \label{tab:main_results}
\end{table*}

We report the results of our experiments in two different ways. We start by first analysing the win-rate of each tested method and then also investigate the quality of the value functions that are learned by all algorithms.

\subsection{Overall performance}
\label{sec:performance}
\begin{figure}[!ht]
\centering
 \includegraphics[width=.95\linewidth]{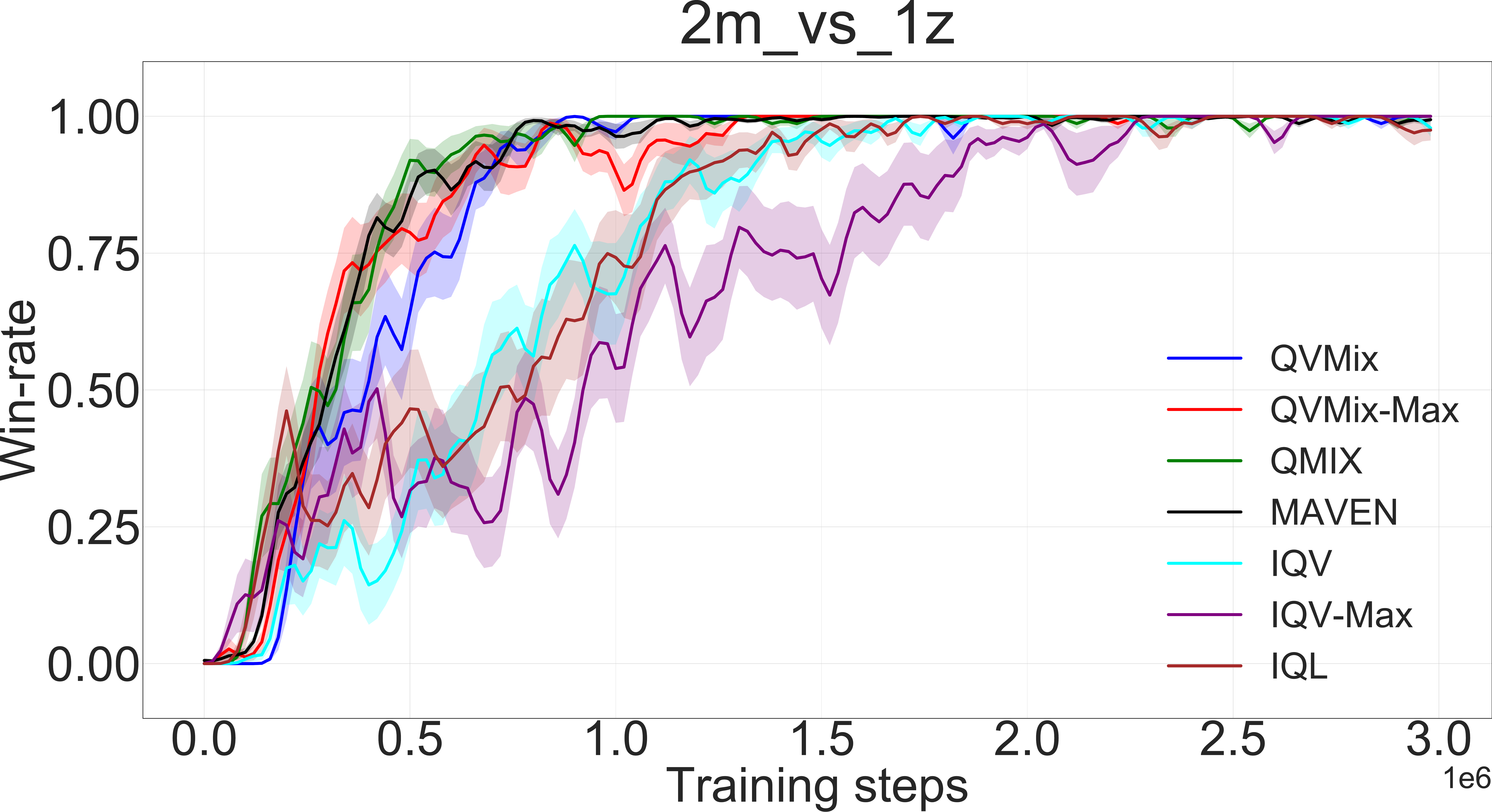}
\caption{Mean of win-rates achieved in the \texttt{2m\_vs\_1z} scenario by QVMix, QVMix-Max, QMIX, MAVEN, IQV, IQVMax and IQL. The error band is proportional to the variance of the measure. We can observe that all methods, although CTDE methods result in faster training than FD methods. We can also see that all four of the novel algorithms based on the DQV family of algorithms can be successfully used in cooperative MARL.}
\label{fig:2m1zwin}
\end{figure}

\begin{figure*}[!h]
\centering
\includegraphics[width=.95\linewidth]{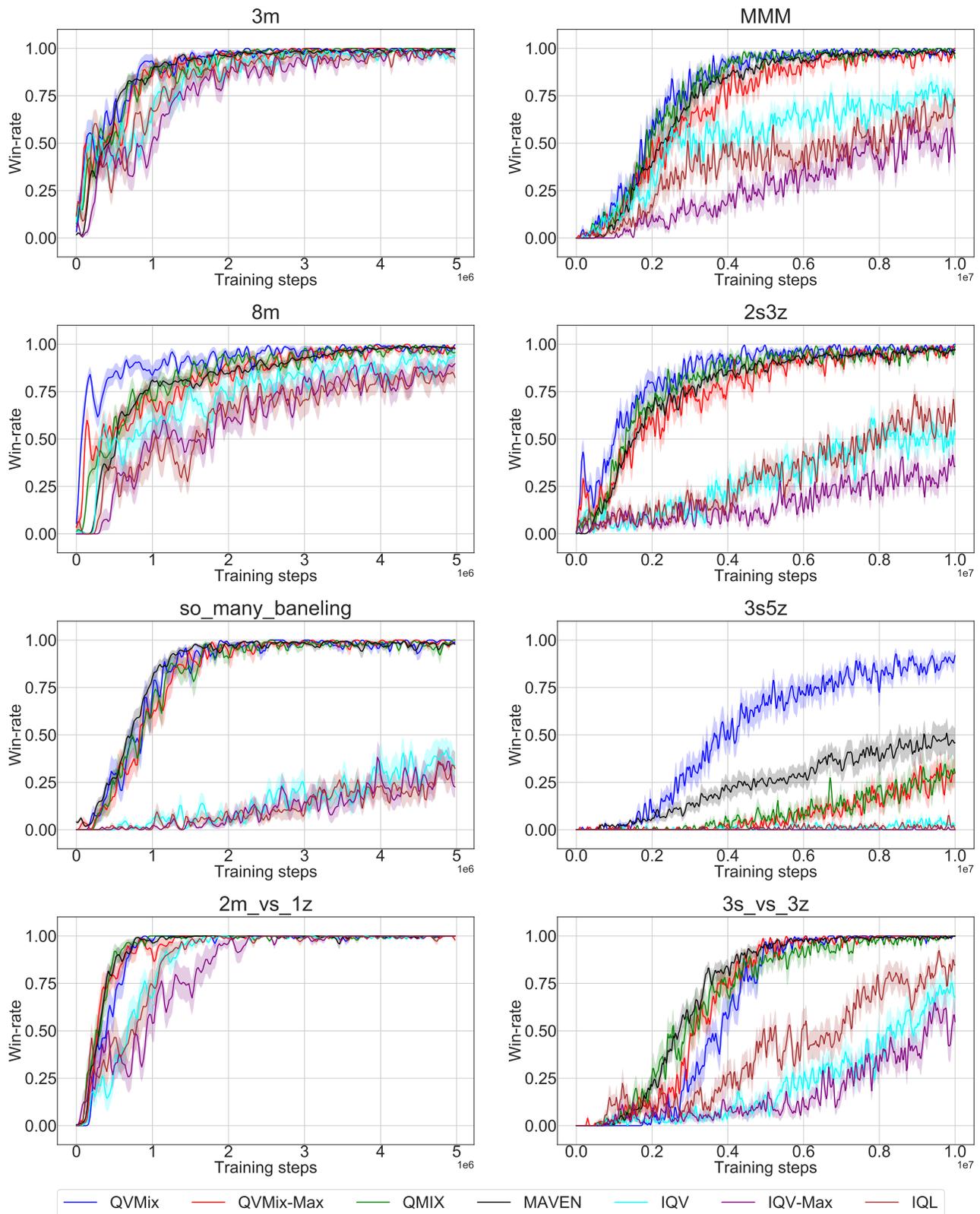}
\caption{Means of win-rates achieved by QVMix, QVMix-Max, QMIX, MAVEN, IQV, IQVMax and IQL in eight scenarios. Top to bottom, left to right, the scenarios are \texttt{3m}, \texttt{8m}, \texttt{so\_many\_baneling}, \texttt{2m\_vs\_1z}, \texttt{MMM}, \texttt{2s3z}, \texttt{3s5z} and \texttt{3s\_vs\_3z}. The error band is proportional to the variance of win-rates.}
\label{fig:all_win_curves}
\end{figure*}

In Table \ref{tab:main_results}, we present for each map and algorithm the respective win-rate
by reporting the means that are measured at the end of training. If the algorithms perform equally in terms of overall performance, meaning that the average win-rate is the same, we consider the one which significantly converges the fastest to be the best performing algorithm. Please note that reporting the win-rate of an episode is a good indicator of the quality of an agent's learned policy, since as introduced in the previous section, a win directly corresponds to the best achievable sum of rewards an agent can receive. We start by observing the differences in terms of performance between the FD methods (IQL, IQV and IQV-Max) and their respective CTDE extensions (QMIX, QVMix and QVMix-Max) and MAVEN.
As one might expect, we can see from the results reported in Fig. \ref{fig:2m1zwin} that FD methods converge more slowly when compared to their CTDE counterparts on the considered \texttt{2\_vs\_1z} map.
This is particularly interesting since it already shows that the DQV family of algorithms can be successfully adapted when it comes to MARL, both in an FD training set-up and in a CTDE one.
However, these results are challenged once the number of agents in the maps increases: examples of such maps are \texttt{so\_many\_baneling}, \texttt{MMM} or \texttt{3s5z}. As is clearly reported both in Table \ref{tab:main_results} and in the plots represented in Fig. \ref{fig:all_win_curves} the performance of FD methods starts to drop, highlighting that CTDE methods are required once the complexity of the training scenario increases.

Therefore, we direct our attention to CTDE methods only and we can then start to see that both
QVMix and QVMix-Max in most of the eight maps perform as well as QMIX and MAVEN.
We again report the evolution of the win-rate of each algorithm on every tested map in Fig. \ref{fig:all_win_curves}. When we consider the \texttt{MMM}, \texttt{3m}, \texttt{2m\_vs\_1z} and the \texttt{so\_many\_baneling} maps we can observe that there is no significant difference between the performance that is obtained by our algorithms and that of QMIX and MAVEN. All methods converge towards the best possible winning-rate and in terms of convergence speed perform closely.

However, when considering the \texttt{2s3z}, \texttt{3s5z} and \texttt{8m} maps we can now observe that the performance of QVMix results in even faster learning. Of even greater interest, when looking at the results obtained on the \texttt{3s5z} map, QVMix is the only algorithm which approaches the best possible win-rate. It is also worth noting that the performance of QVMix-Max is always competitive with the one obtained by QVMix, QMIX and MAVEN. These results do not come as a surprise since similar performance has been observed when DQV-Max was tested in an SARL set-up \cite{sabatelli2020deep}.

\subsection{Overestimation bias}
\label{sec:overestimation_bias}

\begin{figure*}[!h]
\centering
\includegraphics[width=.95\linewidth]{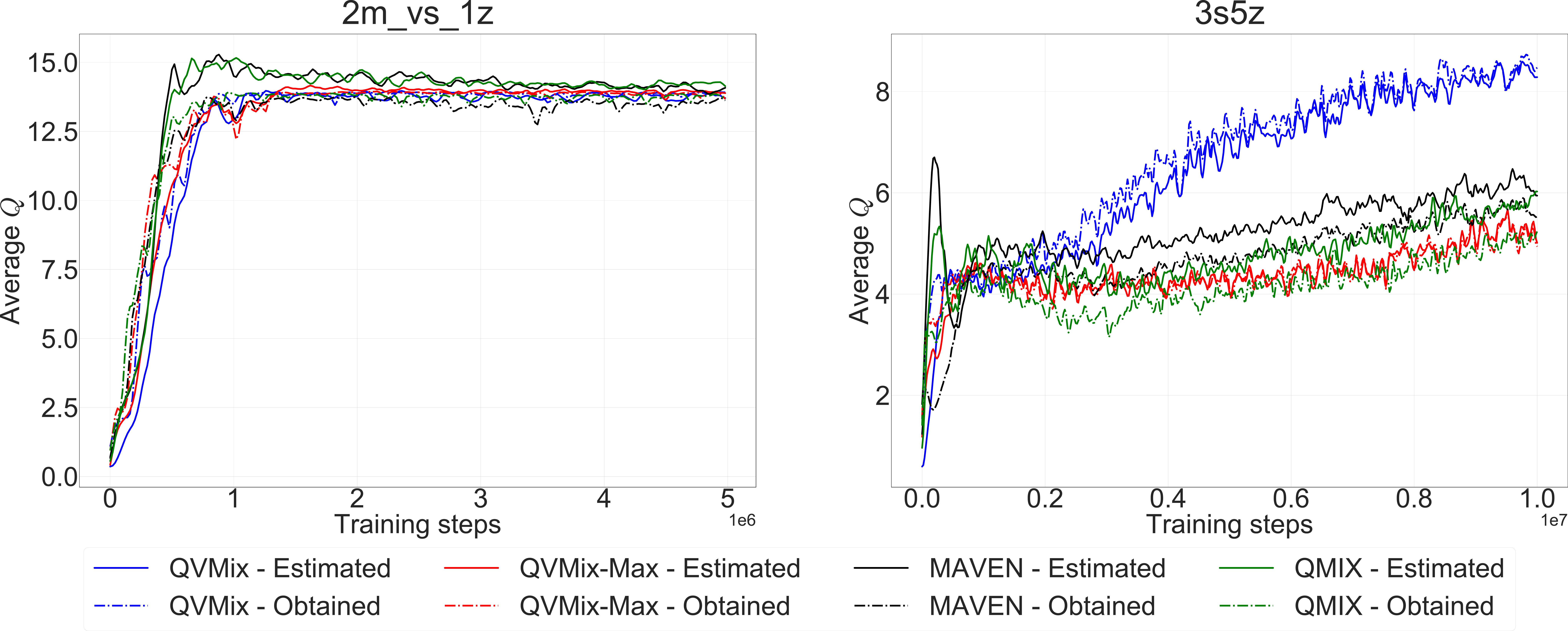}
\caption{$Q$ values obtained and estimed when training QVMix, QVMix-Max, MAVEN and QMIX. Dash-dotted lines represent the obtained $Q$ values while solid lines represent the estimated ones.}
\label{fig:exp_plots_overestim:q_best_worse}
\end{figure*}

To understand the reasons behind why QVMix is the best performing algorithm overall, we analyse how well each method estimates the state-joint-action value function $Q(s_t, \mathbf{}{u_t})$. Since, in most maps, FD methods do not perform as well as CTDE methods, we restrict our analysis to CTDE algorithms only where their respective mixer networks give the estimated $Q(s_t, \mathbf{u_t})$. Since the state space of the maps provided by the SMAC environment is not finite, it is unfortunately not possible to compute the real $Q(s_t, \mathbf{u_t})$ of each state. To overcome this problem, we instead compute the discounted sum of rewards obtained with the current policy in each visited state during an episode and compare the results with the value function inferred from the $Q$ values estimated by the mixer network for these states. The closer the estimates are to the real $Q(s_t, \argmax_{\mathbf{u}}(Q(s_t,\mathbf{u})))$, the more accurate the learned value function is.

For this experiment, we selected two different maps: the \texttt{2m\_vs\_1z} map, which corresponds to the map on which the best results have been achieved by all methods at the same time, and the \texttt{3s5z} map, which on the other hand, is the map on which QVMix performed less well. We report in Fig. \ref{fig:exp_plots_overestim:q_best_worse} the averaged estimated $Q$ values, represented by the solid lines, and the actual discounted sum of rewards, represented by the dash-dotted lines. All are computed for each visited state at testing time.
In both scenarios, we can observe that the $Q$ values that are estimated by QMIX and MAVEN suffer from the overestimation bias of the $Q$ function, while this is not the case for QVMix and QVMix-Max.
We, therefore, justify the better quality of QVMix and QVMix-Max policies by a better approximation of the $Q$ functions, although further work will be required to understand this phenomenon more in detail.

\section{Discussion and Conclusion}
\label{sec:discuss_conclusion}
We introduced four new value-based methods that can be used to train a team of agents in a cooperative MARL framework.
Two of our methods are designed for a fully decentralised execution training scheme (IQV and IQVMax) while two are dedicated to centralised-training with decentralised execution (QVMix and QVMix-Max).
We compared our algorithms with three methods taken from the literature and used the StarCraft Multi-Agent Challenge as benchmark. We have shown that QVMix and QVMix-Max achieve the same results as popular state-of-the-art techniques (QMIX and MAVEN) and that on some of the maps, QVMix can result in faster and better learning. We suggest that this better performance can be related to the fact that QVMix seems to suffer less from the overestimation bias of the $Q$ function.
As future work, it would be interesting to analyze each agent's behaviour and study the impact of the value function in the optimisation procedure. Furthermore we also aim at finding the best possible set of hyper-parameters of each algorithm so that their performance can be exploited even further and more fairly.

\section{Acknowledgments}

PL acknowledges the financial support of the Walloon Region in the context of IRIS, a MecaTech Cluster project.

MS acknowledges the financial support of BELSPO, Federal Public Planning Service Science Policy, Belgium, in the context of the BRAIN-be project.

\bibliography{biblio}
\end{document}